\documentclass[lettersize,journal]{IEEEtran}
\usepackage{amsmath,amsfonts}
\usepackage{algorithmic}
\usepackage{algorithm}
\usepackage{array}
\usepackage[caption=false,font=normalsize,labelfont=sf,textfont=sf]{subfig}
\usepackage{textcomp}
\usepackage{stfloats}
\usepackage{url}
\usepackage{verbatim}
\usepackage{graphicx}
\usepackage{cite}
\usepackage{svg}
\begin{document}

% \documentclass[fleqn,10pt]{olplainarticle}
% % Use option lineno for line numbers 

% \usepackage[utf8]{inputenc}
% \usepackage[english]{babel}
% \usepackage{svg}
% \usepackage{amsmath}

% % \usepackage[
% % backend=biber,
% % style=alphabetic,
% % citestyle=numeric
% % ]{biblatex}

% \usepackage{natbib}
% \setcitestyle{square, comma, numbers,sort&compress, super}
% abbrvnat

% \bibliographystyle{abbrvnat}

\title{{Guiding Machine Perception with Psychophysics}}

\author{\IEEEauthorblockN{
Justin Dulay\IEEEauthorrefmark{1},~\IEEEmembership{Student~Member,~IEEE},
Sonia Poltoratski\IEEEauthorrefmark{2},
Till S. Hartmann\IEEEauthorrefmark{2}, 
Samuel E. Anthony\IEEEauthorrefmark{2}, and Walter J. Scheirer\IEEEauthorrefmark{1},~\IEEEmembership{Senior~Member,~IEEE}
} \\
\IEEEauthorblockA{University of Notre Dame\IEEEauthorrefmark{1}\\
Perceptive Automata, Inc.\IEEEauthorrefmark{2}
}
}

\markboth{Under Review at the Proceedings of the IEEE, April~2022}%
{Shell \MakeLowercase{\emph{et al.}}: A Sample Article Using IEEEtran.cls for IEEE Journals}

%\IEEEpubid{0000--0000/00\$00.00~\copyright~2022 IEEE}

\maketitle

\section*{Introduction}
\IEEEPARstart{G}{ustav} Fechner's 1860 delineation of psychophysics, the measurement of sensation in relation to its stimulus, is widely considered to be the advent of modern psychological science. In psychophysics, a researcher parametrically varies some aspects of a stimulus, and measures the resulting changes in a human subject's experience of that stimulus; doing so gives insight to the determining relationship between a sensation and the physical input that evoked it. This approach is used heavily in perceptual domains, including signal detection, threshold measurement, and ideal observer analysis. Scientific fields like vision science have always leaned heavily on the methods and procedures of psychophysics, but there is now growing appreciation of them by machine learning researchers, sparked by widening overlap between biological and artificial perception~\cite{rojas2011automatic, scheirer2014perceptual,escalera2014chalearn,zhang2018agil, grieggs2021measuring}. Machine perception that is guided by behavioral measurements, as opposed to guidance restricted to arbitrarily assigned human labels, has significant potential to fuel further progress in artificial intelligence. 

In essence, psychophysical measurements of human behavior represent a richer source of information for supervised machine learning. What has been missing thus far from algorithms that learn from labeled data is a reflection of the patterns of error (\emph{i.e.}, the difficulty) associated with each data point used at training time. With knowledge of which samples are easy and which are hard, some measure of consistency can be achieved between the model and the human reference point. The true advantage of doing this stems from the human ability to solve perceptual tasks such as object recognition in an astonishingly fast and accurate way~\cite{dicarlo2012does}. Human visual ability developed over millennia with changes in evolutionary genetic predisposition and thousands of hours of ``pre-training'' for object recognition tasks during development. By leveraging a more powerful learning system --- the brain --- it is possible to improve machine learning training in new ways.

In this \textit{Point of View} article, we advocate for an alternative to traditional supervised learning that operationalizes the science of psychophysics.
We can view the choice of a psychophysical measurement type as a hyperparameter, and the psychophysical measurements themselves as additional labels for data points to be used during training. In traditional supervised learning, performance is limited by the arbitrary labels reflecting class membership, which are the only source of information providing guidance on how to treat individual samples during training. Psychophysically-informed supervised learning is a more complete learning pipeline because of the measured behavioral information. This is, in some ways, akin to the idea of regularization. However, regularization is typically not associated with detailed measurements of human behavior attached to individual data points. 

In the rest of this article, we will take a brief tour of psychophysics for machine learning, including the problem space of perception where these ideas apply, as well as the expanding body of work related to psychophysically-informed machine learning. To highlight the feasibility of gathering and using behavioral measurements in a new machine learning domain, we demonstrate how this training regime works in practice with a series of experiments related to handwritten character classification. As we will see, we are just scratching the surface of what can be achieved with this exciting interdisciplinary area of psychophysically-informed machine learning.

\begin{figure*}[ht]
\centering
\includegraphics[width=1\linewidth]{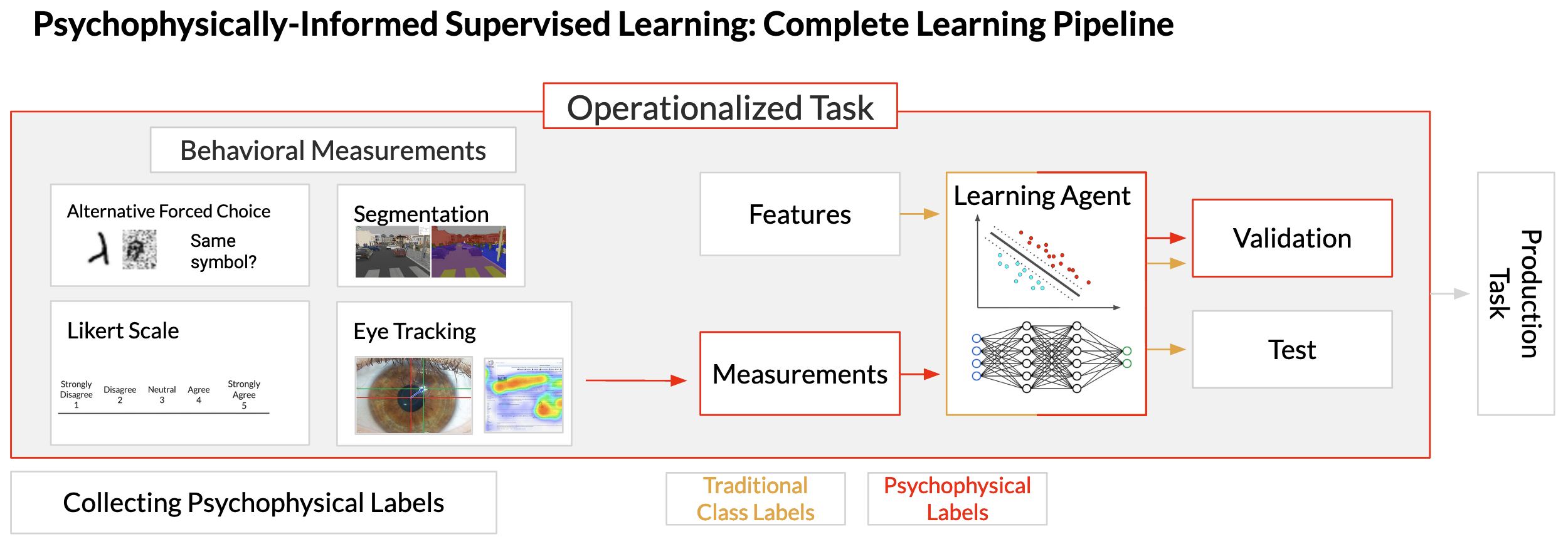}
% \includesvg[width=1\linewidth]{images/key-figure-03.svg}
\caption{When performing similar visual tasks, humans and machine agents both solve for some latent representation of features. But at present, the human capacity for this is superior. The central component of psychophysically informed learning is collecting quantifiable latent information from human experiments on visual recognition tasks and augmenting the \emph{training} regime of machine learning models with it. A learning agent with a closer representational space to humans for a visual task, which is learned from the psychophysical measurements, solves that task in a way that is better than an agent without access to those measurements.}
\label{fig:one}
\end{figure*}

\section*{Psychophysics for Machine Learning}

\textbf{The Problem Space of Perception.}
Hans Moravec's famous AI paradox pointed out that the perceptual tasks which humans accomplish effortlessly have been among the most challenging to model. As he wrote, ``it is comparatively easy to make computers exhibit adult level performance on intelligence tests or playing checkers, and difficult or impossible to give them the skills of a one-year-old when it comes to perception and mobility''~\cite{moravec1988mind}. Indeed, recent advances in computational power have yielded tremendous advances in cognitive tasks that require extensive mental effort for human participants, like advanced strategy games \cite{vinyals2019grandmaster} or machine translation~\cite{wu2016google}. However, machine learning has continued to struggle to perform perceptual tasks that appear intuitive to humans but may have ambiguous or ill-defined ``ground truth,'' like medical image interpretation~\cite{razzak2018deep} or pro-social driving behavior~\cite{lacetera2012will}.

% Indeed, even the most basic building blocks of our visual experience, like color constancy (\cite{foster2011color}) and figure segmentation (\cite{poltoratski2017characterizing}) require complex, nonlinear, and recurrent neural processing of the light signals that enter the eye -- and pose ongoing challenges for machine learning \cite{papernot2016limitations}. Psychophysics has been integral to our understanding of these processes: measuring human perception of just-noticeable-differences (JND, see Glossary) in orientation, for example, can reveal the neural structure of orientation coding in the brain (\cite{wu2017enhanced}). Higher-level visual perception, like the recognition of faces or the categorization of objects, involves an even more complex representational mapping of features to sensory percepts -- which humans often perceive intuitively, without conscious access to rules. A human is able to classify an image of a chair despite being generally unable to express how they do so. They may state that a chair ``should have 4 legs and a sitting surface,'' but can easily realize that many chairs do not fit this definitions while some tables might. Instead of relying on this sort of meta-cognition, psychophysicists intend to resolve the latent space of representational features that yield an observed percept or an appropriate label ('chair' or 'not chair') by measuring reaction time, confusability, and information integration \cite{ashby2005human}. 

We argue that in many cases, both human perceptual science and machine learning intend to resolve the latent space of representational features that yield an observed space. Classification models derive complex latent representations from data without the need for rule-based assignment and can leverage psychophysical data alongside of a class label. For instance, given a photo of a chair, an additional psychophysical measurement associated with the photo would relate information about the latent space of ``chairness'' to complement the extracted features and human assigned label of ``chair.'' This approach is particularly powerful in a machine learning context in which the correct label or solution is conceptually defined by humans, rather than an absolute ground truth. An example of this would be a model of the subjective assignment of ``first impressions'' made about the personality of a face in an image~\cite{rojas2011automatic}.

\textbf{Informing Machine Learning with Psychophysical Data.} The goal of psychophysically-informed data collection is to reveal additional information about the underlying latent representational space that yields the traditional annotations or labels that humans produce for machine learning datasets. This can be done by measuring information about each label's difficulty, confusability of label pairs, or integration of information over time. To cover the latent space effectively, experimental stimuli should effectively cover the sample space of the task, and experiments constructed from the stimuli should span a range of difficulty. Then, experimenters should select a response modality best matched to the machine learning goal and provide careful instructions that focus participant performance on the critical measurement. For example, a task utilizing reaction time would yield more accurate data from keypresses than mouseclicks, and if participants are explicitly instructed to ``perform the task as quickly and accurately as possible, without taking breaks during trials.'' In recent years, psychologists have effectively ported many in-lab study protocols to online crowd-sourcing sites like Amazon Mechanical Turk, demonstrating that data quality can be comparable~\cite{germine2012web} while allowing for rapid data collection from large numbers of participants~\cite{stewart2017crowdsourcing}.

Psychophysical labeling of the data provides an extra dimension beyond the usual supervised label. In a classification mode, a loss function can use this information to improve the learning process. For example, psychophysical labels can be incorporated into the loss to force the learning process to have more consistency with human perception. Consider a loss function where data points with associated low latency in response time result in high error for incorrect model predictions and data points with high latency yield lower error for mistakes. In other words, the model should not make mistakes on easy samples, but is allowed to miss some of the hard ones in the same pattern humans do. Alternatively, there could be some advantage to leveraging the psychophysical information in a way that is \textit{inconsistent} with human behavior, but still improves model performance. One way would be to reverse the error emphasis, so that the training regime puts a higher priority on getting difficult samples correct. 

In a regression mode, Likert-scale data could be used to directly inform training to match human judgement, with additional psychophysical labels for the loss function as needed. For example, a neural network-based regressor can make direct use of averaged Likert scale response scores. As with experimental design in psychology, the options for modeling here are numerous.

%Immediate answers to problems with the same difficulty are more likely to be wrong and are more indicative of a poor-performing or rushing annotator. As practitioners of machine learning, we want to build models that don't simply utilize their bias alone to solve a problem; we want gather as much data to adjust the bias as time (epochs) increase so that we don't overfit to a dataset. 

% say something about choosing the right variable for the right task ...

\begin{figure*}[ht]
\centering
\includegraphics[width=1\textwidth]{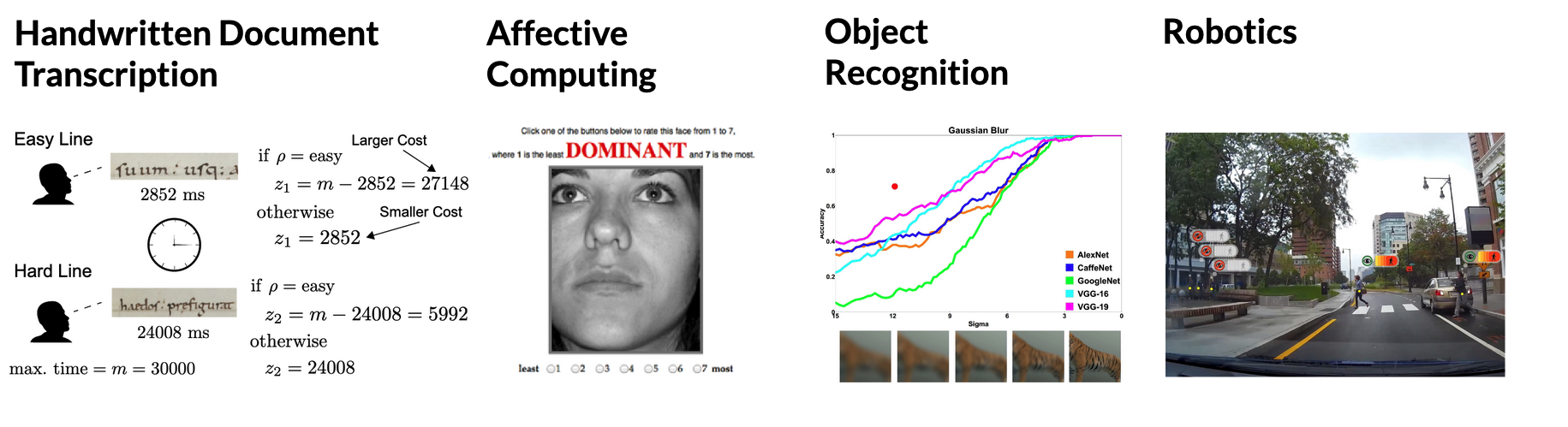}
\caption{Psychophysical labels have been utilized in multiple domains. Supervised model training for Handwritten Document Transcription benefits from measurements of human reading behavior collected via crowd-sourcing site workers or expert readers~\cite{grieggs2021measuring}. Affective Computing tasks can leverage forms of annotation like Likert-scale ratings~\cite{mccurrie2017predicting}. Machine learning algorithms for Object Recognition also benefit from psychophysical evaluation and conditioning~\cite{richardwebster2018visual,zhang2018agil}. The field of robotics, especially pertaining to autonomous vehicles, benefits from labels derived from behavioral measurements when considering social settings like  pedestrian crossings~\cite{milford2019self}. In general, when considering a perceptual task, a supervised learning model can always benefit from a larger label space.}
\label{fig:two}
\end{figure*}

\textbf{Domains that Have Benefited from This Approach.} Several domains have been investigated by researchers looking for ways to use psychophysics in machine learning (Fig.~\ref{fig:two}). A loss function able to use measurements from crowd-sourced psychophysics experiments was first introduced by Scheirer et al.~\cite{scheirer2014perceptual} for the domain of human biometrics. They conducted a series of behavioral experiments using the psychology crowd-sourcing platform \texttt{TestMyBrain.org} where participants were presented with a two-alternative forced choice question about whether or not a face was present in a given stimulus. Two-alternative forced choice is a common way to gather psychophysical data for recognition-based tasks because the operative step of recognition is binary --- selecting a given positive sample type in a relative latent representation space and rejecting a given negative sample~\cite{prins2016psychophysics}. Stimuli were designed to test the impact of different controllable conditions such as noise and occlusion on face detection, which provides increased coverage of the full difficulty and scope of the problem compared to unperturbed labeling, maximizing the size of an informative label space. Different from the typical treatment of labeled samples in supervised machine learning, they found that difficult samples (\emph{e.g.}, a heavily occluded face) where the participants chose correctly after a relatively long period provided additional information for training Support Vector Machine (SVM) classifiers with a loss function applying penalties based on perceptual measurements. Adding psychophysical measurements increased the robustness of the \emph{label space}, which led to a state-of-the-art model for face detection.

%%% Sam's paper, which follows from the SVM work.
In the domain of handwritten document transcription, loss functions incorporating psychophysical data for artificial neural network training have also been explored. Grieggs et al.~\cite{grieggs2021measuring} measured the reaction time of expert readers for documents of varying age and language, and used those measurements as labels in different loss functions that could emphasize easy or difficult samples. Observations were made about differences in reader behavior between expert and novice groups, with implications for other data labeling tasks. As with the original SVM work, this strategy was able to yield state-of-the-art performance for handwritten document transcription.

% This sentence above needs some help. Reaction time annotations helped the loss function? I don't think reaction time is helping. Maybe: adding reaction times to the loss function stilled allowed the model to generalize. Though I don't understand the still here. (tried to address this - justin)

%%% Likert papers. Our stuff + ChaLearn
It has been demonstrated that humans adeptly make complex judgements about personality traits in miniscule amounts of time~\cite{willis2006first}. In the domain of affective computing, using two-alternative forced choice or Likert scale ratings and regression models, it is possible to model this phenomenon using machine learning. The ChaLearn Looking at People First Impressions Challenge  Competition~\cite{ponce2016chalearn} focused on models for the Big 5 personality traits (Openness, Conscientiousness, Extraversion, Agreeableness, and Neuroticism). McCurrie et al.~\cite{mccurrie2017predicting} looked at a set of different traits, including Trustworthiness, Dominance, IQ and Estimated Age. Work by Rojas et al. has looked at modeling some of these traits in a classification mode~\cite{rojas2011automatic}. This research is a natural extension of laboratory testing in social psychology, and is facilitated by psychophysical measurements. 

%%% Psychophysics for evaluation. DiCarlo. Bethge. Others.
% \cite{dicarlo2012does}
% \cite{geirhos2018generalisation} 
% Domain of object recognition. Other variations on psychophysics and machine learning also exist in the literature. 
In the domain of object recognition, an agent attempts disentangle a learned manifold on some latent representation space of learned data~\cite{dicarlo2012does}.
Psychophysical evaluation has played an important role in evaluating biological and artificial vision, including their similarities and differences. Perturbed stimuli (\textit{e.g.}, rotated objects) activate neurons at different levels than canonical object views of the same stimuli. This same effect was observed in artificial neural networks~\cite{webster2018psyphy}. But there are also differences between models and biological reference points. For instance, while some artificial neural networks generalize better than humans on some types of noise, humans outperform them on many noisy recognition tasks ~\cite{behavioralgeirhos2018generalisation, jang2021noise}. This work demonstrates that the incorporation of human behavioral measurements within the label space of specific recognition tasks where the artificial agent typically fairs poorly can be beneficial. RichardWebster et al. introduced a framework to evaluate different types of image perturbations (blurring, rotation, resolution) and their effects on both artificial and human performance for object~\cite{webster2018psyphy} and face recognition~\cite{richardwebster2018visual} tasks. Zhang et al.~\cite{zhang2018agil} have suggested the use of human gaze measurements for improving performance in various object-related tasks, especially in a reinforcement learning context.

%%% Robotics. 
Finally, in the domain of robotics, psychophysics has been positioned as a means to create more generalizable embodied intelligence in robotic systems \cite{sunderhauf2018limits}. By simulating an artificial environment with many potential scenarios using perturbed inputs, a robotic system can generalize by learning to perform on stimuli that could potentially appear in the wild. This borrows from human-in-the-loop learning for autonomous systems, but remains distinct in that the measurements for the scenarios are taken from people before the model is trained. Furthermore, it has also been suggested that the models used for autonomous driving can benefit from the human perception of pedestrians in uncertain situations (\textit{e.g.}, a pedestrian at a crosswalk who is not in motion, but may intend to cross), which reveals more information about the situation at hand~\cite{milford2019self}.

\section*{Case Study: Optical Character Recognition with Psychophysical Parameterization}

% - describe ML task 
% - all the dataset and experiment work right away 
% - loss function (w differences between RT and acc beforehand
% - experimental implementation

Optical character recognition (OCR) is a popular supervised learning task where the objective is to classify the characters within images of text. Sometimes those images are of poor quality, rendering the task more challenging to the learning agent while providing a representative example of text in the wild. Humans, equipped with a rich latent understanding of text recognition, typically outperform artificial agents on nontrivial OCR tasks. In this case study, we conducted a series of human behavioral measurements in crowd-sourced experiments to operationalize the training stage of a supervised deep learning agent with psychophysical data.

% To demonstrate how straightforward it is to inform a machine learning algorithm with psychophysical data, we will walk through an experiment related to Optical Character Recognition with characters that are novel to human observers --- a new domain for this approach. We demonstrate that a modest crowd-sourced data collection can yield high-quality psychophysics data, which can then be used to create a robust representation space for the learning agent. Augmenting a labeled data set of  characters with recorded human reaction time measurements yielded substantive increases in accuracy for artificial neural networks trained with a loss function that can use the psychophysical data.

% dataset section flow 
% - describe what omniglot is, and why we chose it for this task (vs mnist and others, unfamiliar symbols)
% - describe how we partitioned the dataset, split via gan
% - describe the psychophysical task
% - describe the different experiments, from the control to the rewording to the perturbations
% - describe how the task can augment model (with reaction time and accuracy), further refernce to loss

\textbf{Dataset Preparation and Behavioral Experiments.} 
We implemented both the psychophysical tasks and the OCR machine learning task using a subset of the Omniglot dataset~\cite{lake2015human}. The dataset contains images of handwritten characters from hundreds of typesets, many of which a typical crowd-sourced study participant would be unfamiliar with. In order to prepare a stimulus dataset for the human behavioral experiments, we selected 100 random classes from the original Omniglot dataset. To augment this data, we generated a counterpart sample for each image with a deep convolutional generative adversarial network (DCGAN) to increase intraclass variance and the sample size per class. This resulted in a dataset of 100 classes with 40 instances per class for the psychophysical stimulus dataset to be used by the participants. 

% tasks
We conducted a series of four different psychophysical behavioral experiments on variations of a two-alternative forced choice task with human participants. For each experiment of this particular task, the participant viewed two different images from the stimulus dataset with a prompt asking whether the images represent the same symbol or not. The first image of the pair was chosen at random, while the second was chosen from the same class or a different one with a probability of 0.5.

\begin{figure}[t]
\centering
\includegraphics[width=1\linewidth]{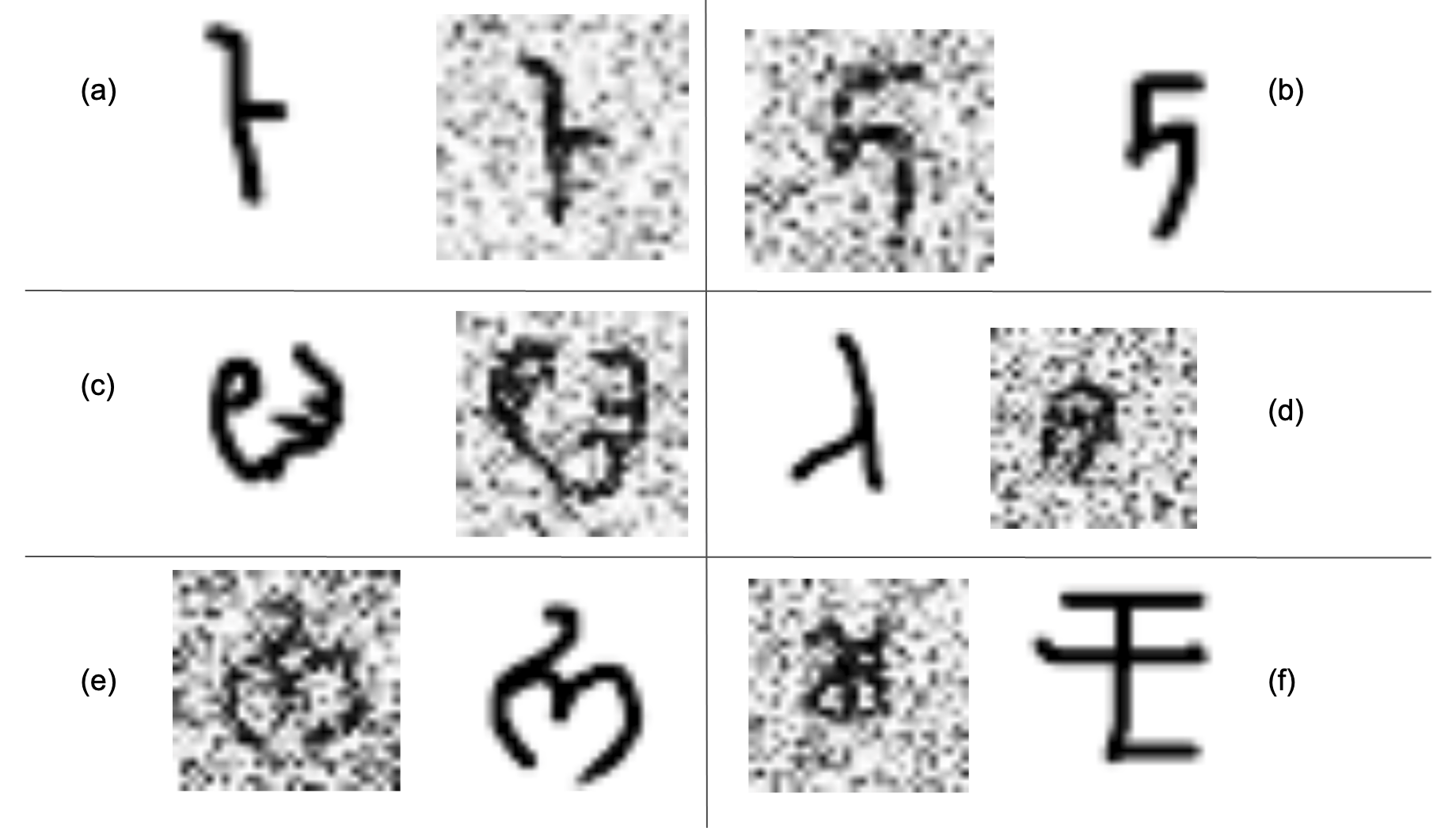}
\caption{\emph{Are these the same character?} An example two-alternative forced choice OCR task as seen from the participant's view. Labels (d) and (f) represent character pairs where the class labels differ; the rest represent the same class pairing. 
% In the tasks which used the Gaussian noise to perturb the inputs, we weighted most tasks with a higher probability of moderate noise. 
The blurred and noisy images lead to more informative psychophysical labels for operationalization within the machine learning task during training.}
% The goal of this was to create a task space that challenges the subject --- if the task is too easy, they tend resort to overconfident guessing; if the task is too hard, they tend to give up on the task. The Gaussian noise perturbation was the most effective at increasing classification accuracy when psychophysically informing its feature representation space.}
\label{fig:three}
\end{figure}

% notes on making the commented paragraph more clear to the reader
\begin{itemize}
    \item The first experiment was a control experiment where the images from the original stimulus dataset were not perturbed.  As a baseline, this task presented instructional prompts that were not tailored to psychophysics tasks, but rather asked participants for standard machine learning labels. Users made their responses using a cursor, which is typical in labeling tasks but does not yield reliable reaction-time estimates.
    \item The second experiment was the same as the control experiment, but we modified the instructional prompts following best practices in psychophysics --- for example, participants were instructed to complete the task ``as quickly and accurately as possible'' and they were allowed to complete the task by pressing an \emph{F} or \emph{J} key. Both of these modifications are standard in psychophysics tasks that collect response time data. 
    % Additional wording changes were made to clarify the task (see Appendix).
    \item The third experiment incorporated the condition of Gaussian blur. A randomly chosen image from one of the 100 classes was blurred using one of five different kernels (also chosen randomly with respect to the level of perturbation), and the other image that was paired with it was left unaltered. 
    We expanded the range of experiment difficultly to avoid a ceiling effect, a form of scale attenuation in which maximum performance measured does not reflect the true maximum of the independent variable. In this case, we expect a measurement ceiling if the task was too easy for participants and maximally-accurate responses lose their relationship to task difficulty.
    % This was done to expand the range of difficulty across task experiments, allowing us to sample across a wider span of human performance on this task and avoiding ceiling effects (CITE).
    \item The fourth experiment was conducted like the third experiment, but with Gaussian noise instead of blurring. Likewise, there were five different levels of Gaussian noise that could be applied, selected at random. Refer to Fig.~\ref{fig:three} for sample depictions of this task. 
\end{itemize}

% The (1) first experiment was a control, where the images and prompts were not modified from the original stimulus dataset. We believed the task prompts could have been more succinct, and we modified them in \emph{subsequent} experiments. The (2) second experiment was the same as the control, but we modified the prompts for clarity --- for example, participants were instructed to complete the task ``as quickly and accurately as possible'' and they were allowed to complete the task by pressing an \emph{F} or \emph{J} key. Next, we explored how perturbing the images would change the psychophysical data for each response. The (3) third experiment was also a two-alternative forced choice task, but incorporated the condition of Gaussian blur. A randomly chosen image from one of the 100 classes was blurred using one of five different kernels (also chosen randomly with respect to the level of perturbation), and the other image that was paired with it was left the same. The (4) fourth experiment was conducted like the third experiment, but with Gaussian noise instead of blurring. Likewise, there were five different levels of Gaussian noise that could be applied, selected at random. Refer to Fig.~\ref{fig:three} for sample depictions of this task.

% participants 
Participants for the behavioral experiments were recruited on Amazon Mechanical Turk. Each participant completed 100 two-alternative forced choice trials, and each of the four experiments had 1000 participants. The reaction time for each task was recorded by measuring the interval between the first presentation of the stimuli and the participant's recorded response. Each experiment formed a psychophysically annotated dataset that was used later in the machine learning task. The resulting four datasets included all of the image pairs shown in each two-alternative forced choice instance, the responses of the participants, the average accuracy of participants on each image pairing, and the average reaction time of participants on each image pairing. Each image pairing was distributed approximately evenly across all participants in each experiment. The averaged accuracies and reaction times per pairing were calculated across all responses for that instance, where the number of responses per pairing varied slightly but not significantly. Spam and incomplete response sets were manually pruned.

\textbf{Loss Function Formulation.} 
The psychophysical loss utilized data collected from the human behavioral experiments in addition to traditional supervised learning data. It expanded upon a traditional supervised learning pipeline. In this case study, we made use of a standard ResNet50 deep neural network model, which is ubiquitous in many computer vision classification tasks, and cross entropy loss. For all experiments, we used the same hyperparameters for the model. 

% Here we describe the loss function utilized during training. In our data collection, we gathered the reaction time associated with each image pair and computed the average accuracy score per label. We chose to parameterize cross entropy loss, the most common loss function for multi-class classification tasks. The loss function computes the log error of the softmax outputs of each class. The advantage of using a cross entropy loss function in a multiclass classification task is that it assigns a weighted probability for each weighted class and activates on the most likely option --- not unlike the human within an alternative forced choice task. 

% The strategy used here is to penalize wrong predictions on short reaction times more severely than ones with longer reaction times. A HIT with a short reaction time aligns with the hypothesis that the question should be easier. We calculate this by assigning a penalty $z$ at moment $i$, where $i$ is the current class index:

Cross entropy loss is defined as: 
$$\mathcal{L} = -\left(\sum_jy_jlog(\hat{y}_j)\right)$$  
where $\hat{y}_j$ is a model prediction and $y_j$ is the traditional class label associated with it. In order to incorporate the psychophysical labels into cross entropy loss, we normalized and scaled the measurements to fit within the expected range of the loss function values. Further, we only considered modifying the behavior of the loss function on model outputs where the prediction was incorrect. We made use of the averaged reaction times and averaged accuracies separately from one another; we did not combine the two in a given loss function. To make use of these labels, we defined a psychophysical penalty:
$$z_i = m - r_i$$ where $z_i$ is the penalty, $m$ is the maximum value for either reaction time or  accuracy, and $r_i$ is the psychophysical label (either reaction time or accuracy). Next, we incorporate $z_i$ into cross entropy loss: 
$$\mathcal{L} = -\left(\sum_jy_j\left(log(\hat{y}_j) z_ic)\right)\right)$$ where $c$ is a scaling factor for the psychophysical penalty. 

% Initially, parameterizing cross entropy loss with reaction time caused a scaling issue. If the loss function inputted the raw values of the latency times from the dataset, the loss function would infinitely span towards a zero asymptote, resulting in \texttt{nan} values. This persists even when reporting a scaling to the raw data values themselves. To account for all instances of this while still accounting for the appropriate loss penalty, we add an additional scaling factor before computing the softmax logits. 

% The psychophysical loss function is: 

% $$\mathcal{L} = -\left(\sum_jw_jlog(x_j z_i * c)\right)$$

% where $c$ is a scaling factor for the psychophysical penalty. 

%As observed, we chose the specific instruction wording, psychophysical parameters, and noise and blurring scales to most efficiently extract the right psychophysical parameters for this deep learning system. While other optimizations could have been made, such as dataset refinement or model selection, we chose this experimental setup to isolate the effect of parameterizing ML systems with psychophysical parameters. 

\textbf{OCR Classifier Experiments.}
The study concluded that psychophysical loss improves the top-1 accuracy of the dataset by 1.1\% on average --- a substantial improvement for a machine learning endeavour. 
The ResNet50 architecture used in these experiments was pre-trained on ImageNet. We trained three models based on this architecture for each of the four psychophysical datasets from the behavioral experiments: a set of equally unperturbed images (control experiment), re-worded prompts for the control experiment set, blurred images, and noisy images. 

\begin{itemize}
    \item The first model was a standard ResNet50 with normal cross entropy loss.
    \item The second model substituted regular cross entropy loss for the psychophysical loss using accuracy.
    \item The third model substituted regular cross entropy loss for the psychophysical loss using reaction time.
\end{itemize}

We trained each model for 20 epochs. In order to report accuracy fairly, we repeated model training five times with a different random seed. The results reported in Table~\ref{tab:widgets} reflect the mean accuracy of each run along with standard error. In addition, we conducted a three-way ANOVA between the reaction time, accuracy, and cross-entropy sets in this experiment.

% experimental results steps
% - model formulation
% - experiment descriptions
% - for each psychophysical set 
%     - control 
%     - rt
%     - accuracy

% We found a learning rate of 0.01 to be best overall for 20 epochs. Likewise, we used an ADAM optimizer to efficiently compute gradients \cite{kingma2014adam} during the backward step of the model. We chose this deep learning setup specifically because it is representative of common practice in deep learning. 

% experiment formulation

% Each task was conducted sequentially. First, we dealt with a control group of a task with no augmentations. Next, we repeated the experiment with different wording on all parts of the instructions, subjectively easier to understand. We repeated this with the addition of blurring to one of the images within the task window, but not both. The blur level was randomly selected in a range from no blur to an indistinguishable blob. After this, repeated the blur experiment, but with Gaussian noise instead. As the table suggests, adding the noise image set best augments the machine learning model performance. Also, we posit that the average accuracy per class did not augment the ML model effectively because labels were not always compared with the same matches - accuracy gets much more noisy when comparing different labels as opposed to reaction time. The reaction time best informs the model whenever the questions are closer to a boundary within the latent feature representation space.

% okay let's add the 95% CI for the table
\begin{table}[ht]
\centering
\begin{tabular}{l|l|l|l}
\textcolor{violet}{Control experiment} & \textcolor{violet}{Train Accuracy}  & \textcolor{violet}{Test Accuracy} & \textcolor{violet}{95\% C.I.} \\
Cross Entropy & 0.741 $\pm0.005$ & 0.705 $\pm$0.004 & 0.078 \\
Averaged Accuracy & 0.743 $\pm0.005$ & 0.692 $\pm0.008$ & 0.055 \\
\textbf{Reaction Time} & \textbf{0.754 $\pm$0.005} & \textbf{0.719 $\pm$0.004} & 0.055 \\\hline
\textcolor{red}{Different Prompts} & \textcolor{red}{Train Accuracy} & \textcolor{red}{Test Accuracy} & \textcolor{red}{95\% C.I.} \\ 
Cross Entropy & 0.732 $\pm0.005$ & 0.705 $\pm$0.004 & 0.078 \\
Averaged Accuracy & 0.723 $\pm0.003$ & 0.697 $\pm$0.008 & 0.062\\
\textbf{Reaction Time} & 0.731 $\pm$0.005  & \textbf{0.729 $\pm$0.005} & 0.062 \\\hline
\textcolor{blue}{Blurred Images} & \textcolor{blue}{Train Accuracy} & \textcolor{blue}{Test Accuracy} & \textcolor{blue}{95\% C.I.} \\ 
Cross Entropy & 0.691 $\pm0.005$ & 0.642 $\pm$0.005 & 0.062\\
Averaged Accuracy & 0.643 $\pm0.008$ & 0.542 $\pm$1.005 & 0.107 \\
\textbf{Reaction Time} & \textbf{0.710 $\pm$0.003} & \textbf{0.668 $\pm$0.006} & 0.068 \\ \hline
\textcolor{teal}{Noisy Images} & \textcolor{teal}{Train Accuracy} & \textcolor{teal}{Test Accuracy} & \textcolor{teal}{95\% C.I.} \\ 
Cross Entropy & 0.672 $\pm0.006$ & 0.602 $\pm$0.005 & 0.062\\
Averaged Accuracy & 0.641 $\pm0.007$ & 0.592 $\pm$0.016 & 0.110 \\
\textbf{Reaction Time} & \textbf{0.732 $\pm$0.004} & \textbf{0.680 $\pm$0.005} & 0.062
\end{tabular}
\caption{\label{tab:widgets} The test time accuracy for Top@1 accuracy reflects substantial benefit in using reaction time as an additional label. The results have a p-value from ANOVA of $2.02\cdot10^{-6}$, indicating significant difference in the performances when using reaction time as a psychophysical label in the model. Likewise, the confidence intervals for each set of experiments remains within expected bounds of efficacy for consistent performance across folds. Using the averaged accuracy score as a label rarely yielded substantial benefit. However, we see improved performance when using reaction time as a psychophysical parameter in all cases.}
\end{table}

% \begin{table}[ht]
% \centering
% \begin{tabular}{l|r}
% Top@1 Accuracy on Test Data & Control\\\hline
% Cross Entropy & 0.705 $\pm$0.4\\
% Averaged Accuracy Parameter & 0.692 $\pm0$.8 \\
% \textbf{Reaction Time} & \textbf{0.719 $\pm$0.4} \\\hline
% Top@1 Accuracy & Different Prompts \\ \hline
% Cross Entropy & 0.705 $\pm$0.4\\
% Averaged Accuracy Parameter & 0.697 $\pm$0.8\\
% \textbf{Reaction Time} & \textbf{0.769 $\pm$0.5} \\\hline
% Top@1 Accuracy &  Blurred Images \\ \hline
% Cross Entropy & 0.642 $\pm$0.5\\
% Averaged Accuracy Parameter & 0.542 $\pm$1.5\\
% \textbf{Reaction Time} & \textbf{0.668 $\pm$0.6} \\ \hline
% Top@1 Accuracy & Noisy Images \\ \hline
% Cross Entropy & 0.602 $\pm$0.5\\
% Averaged Accuracy Parameter & 0.592 $\pm$1.6\\
% \textbf{Reaction Time} & \textbf{0.680 $\pm$0.5} 
% \end{tabular}
% \caption{\label{tab:widgets} The test time accuracy for Top@1 accuracy reflects substantial benefit in using reaction time as a parameter. This remains consistent across several hyperparameter tunings and model changes but remains most pronounced here. Likewise, parameterizing the averaged accuracy score rarely yielded substantial benefit. These results support that adding reaction time labels alone in the parameter space improve classification tasks. Learning to differentiate among easy and hard tasks with labels provides additional benefit as opposed to feature-based learning alone. Also, we utilize five different random seeds for each experiment run for brevity.}
% \end{table}

 For each instance, reaction time labels generally improved supervised model performance. In contrast, accuracy labels did not always outperform the control. Therefore, when integrating these new labels into machine learning training, it remains important to assess effectiveness for the task. In this case, reaction time was the more informative measurement type. This has been shown in the literature for training artificial neural networks using psychophysical data~\cite{grieggs2021measuring}. However, there is not guarantee that this will generalize to all tasks. A two-way ANOVA between the reaction time and control test result distributions established statistical significance.

%  The ANOVA test established the statistical significance of the effect when using reaction time with the psychophysical loss. 
 
% This not conclude that reaction time is always better than average accuracy for parameterization of label spaces; rather, it demonstrates that for the ML task, a psychophysical annotation outperforms the control significantly. This example experiment demonstrates that while not all psychophysical parameterizations improved the performance of learning tasks, the leveraging of the human cognition within the machine learning process generally leads to greater model performance improvements with added with care. 

\section*{Conclusion} 

% Our experiment demonstrated a novel improvement in the state-of-the-art performance of a deep learning model on optical character recognition using psychophysical parameters.
% In this case, reaction time worked as the best choice parameter to improve the learning agent; however, it may not always work. This brings us to the next point: the particular psychophysics parameter depends upon the specific task, but the general utilization of psychophysical parameters in a human knowledge related task domain generally improves learning model performance. This experiment, while specific in nature, demonstrates that there may be many possible use cases for psychophysical parameterization of models in the future. Likewise, as long as the machine learning task attempts to solve a latent feature space representation of a task that is similar to what a human solves, the usage of these latent parameters will likely be beneficial. 

Psychophysical labels from human behavioral experiments have been shown to improve the performance of supervised learning models in many different domains in the literature. We conducted a case study to demonstrate how quickly this strategy can be adapted to a new domain. More work needs to be done to develop similar strategies for different modes of learning, including unsupervised and reinforcement learning. By improving training regimes or policy estimators in these fields, generalization may be achieved more effectively than with traditional strategies. In all, we hope this work inspires future conversation and research at the intersection of psychology and computer science. 

% Psychophysics remains a demonstrably strong and depth-laden field to apply to machine learning. The richness of past experimentation, the cheapness of HIT tasks, and the measurable effectiveness of the results contribute to the case for usage within the deep learning community. The two strike a delicate complementary balance of human and growth over time. Deep learning has markedly improved in its capacity to memorize and store useful features. This is akin to an original computer exponentially increasing in its memory abilities or how the human mind can hold vast amounts of raw information. The intuition, and even the beauty of the psychophysics data, adds something inherently human to to the continuous decision making process of a deep learning neural network. As these networks grow increasingly complex, and increasingly more challenging to interpret, the need for explainability grows. What better way to explain a black box than to introduce human data, human thinking and reaction, to the parameter space itself? 

\section*{Acknowledgment}
This work was partially funded by the US National Science
Foundation under grant BCS:1942151.

% \section*{Declaration of Interests}
% J.D. and W.J.S. declare no conflict of interests. S.P., T.S.H., and S.E.A work for Perceptive Automata, a company selling psychophysics based machine learning models. 

\section*{Data and Code Availability}
All data and code used in this paper can be found at: https://github.com/dulayjm/PyTorch-Psychophysics-Learning.

% the glossary and questions are now in separate files
\bibliography{sample.bib}

\begin{thebibliography}{10}
\providecommand{\url}[1]{#1}
\csname url@samestyle\endcsname
\providecommand{\newblock}{\relax}
\providecommand{\bibinfo}[2]{#2}
\providecommand{\BIBentrySTDinterwordspacing}{\spaceskip=0pt\relax}
\providecommand{\BIBentryALTinterwordstretchfactor}{4}
\providecommand{\BIBentryALTinterwordspacing}{\spaceskip=\fontdimen2\font plus
\BIBentryALTinterwordstretchfactor\fontdimen3\font minus
  \fontdimen4\font\relax}
\providecommand{\BIBforeignlanguage}[2]{{%
\expandafter\ifx\csname l@#1\endcsname\relax
\typeout{** WARNING: IEEEtran.bst: No hyphenation pattern has been}%
\typeout{** loaded for the language `#1'. Using the pattern for}%
\typeout{** the default language instead.}%
\else
\language=\csname l@#1\endcsname
\fi
#2}}
\providecommand{\BIBdecl}{\relax}
\BIBdecl

\bibitem{rojas2011automatic}
M.~Rojas~Q., D.~Masip, A.~Todorov, and J.~Vitria, ``Automatic prediction of
  facial trait judgments: Appearance vs. structural models,'' \emph{{PLOS
  ONE}}, vol.~6, no.~8, p. e23323, 2011.

\bibitem{scheirer2014perceptual}
W.~J. Scheirer, S.~E. Anthony, K.~Nakayama, and D.~D. Cox, ``Perceptual
  annotation: Measuring human vision to improve computer vision,'' \emph{IEEE
  Transactions on Pattern Analysis and Machine Intelligence}, vol.~36, no.~8,
  pp. 1679--1686, 2014.

\bibitem{escalera2014chalearn}
S.~Escalera, X.~Bar{\'o}, J.~Gonzalez, M.~A. Bautista, M.~Madadi, M.~Reyes,
  V.~Ponce-L{\'o}pez, H.~J. Escalante, J.~Shotton, and I.~Guyon, ``Cha{L}earn
  looking at people challenge 2014: Dataset and results,'' in \emph{ECCV
  Workshops}, 2014, pp. 459--473.

\bibitem{zhang2018agil}
R.~Zhang, Z.~Liu, L.~Zhang, J.~A. Whritner, K.~S. Muller, M.~M. Hayhoe, and
  D.~H. Ballard, ``{AGIL}: Learning attention from human for visuomotor
  tasks,'' in \emph{ECCV}, 2018, pp. 663--679.

\bibitem{grieggs2021measuring}
S.~Grieggs, B.~Shen, G.~Rauch, P.~Li, J.~Ma, D.~Chiang, B.~Price, and
  W.~Scheirer, ``Measuring human perception to improve handwritten document
  transcription,'' \emph{IEEE Transactions on Pattern Analysis and Machine
  Intelligence}, 2021.

\bibitem{dicarlo2012does}
J.~J. DiCarlo, D.~Zoccolan, and N.~C. Rust, ``How does the brain solve visual
  object recognition?'' \emph{Neuron}, vol.~73, no.~3, pp. 415--434, 2012.

\bibitem{moravec1988mind}
H.~Moravec, \emph{Mind Children: The Future of Robot and Human
  Intelligence}.\hskip 1em plus 0.5em minus 0.4em\relax Harvard University
  Press, 1988.

\bibitem{vinyals2019grandmaster}
O.~Vinyals, I.~Babuschkin, W.~M. Czarnecki, M.~Mathieu, A.~Dudzik, J.~Chung,
  D.~H. Choi, R.~Powell, T.~Ewalds, P.~Georgiev \emph{et~al.}, ``Grandmaster
  level in {StarCraft II} using multi-agent reinforcement learning,''
  \emph{Nature}, vol. 575, no. 7782, pp. 350--354, 2019.

\bibitem{wu2016google}
Y.~Wu, M.~Schuster, Z.~Chen, Q.~V. Le, M.~Norouzi, W.~Macherey, M.~Krikun,
  Y.~Cao, Q.~Gao, K.~Macherey \emph{et~al.}, ``Google's neural machine
  translation system: Bridging the gap between human and machine translation,''
  \emph{arXiv preprint arXiv:1609.08144}, 2016.

\bibitem{razzak2018deep}
M.~I. Razzak, S.~Naz, and A.~Zaib, ``Deep learning for medical image
  processing: Overview, challenges and the future,'' \emph{Classification in
  BioApps}, pp. 323--350, 2018.

\bibitem{lacetera2012will}
N.~Lacetera, M.~Macis, and R.~Slonim, ``Will there be blood? {Incentives} and
  displacement effects in pro-social behavior,'' \emph{American Economic
  Journal: Economic Policy}, vol.~4, no.~1, pp. 186--223, 2012.

\bibitem{germine2012web}
L.~Germine, K.~Nakayama, B.~C. Duchaine, C.~F. Chabris, G.~Chatterjee, and
  J.~B. Wilmer, ``Is the web as good as the lab? {Comparable} performance from
  web and lab in cognitive/perceptual experiments,'' \emph{Psychonomic Bulletin
  \& Review}, vol.~19, no.~5, pp. 847--857, 2012.

\bibitem{stewart2017crowdsourcing}
N.~Stewart, J.~Chandler, and G.~Paolacci, ``Crowdsourcing samples in cognitive
  science,'' \emph{Trends in Cognitive Sciences}, vol.~21, no.~10, pp.
  736--748, 2017.

\bibitem{mccurrie2017predicting}
M.~McCurrie, F.~Beletti, L.~Parzianello, A.~Westendorp, S.~Anthony, and W.~J.
  Scheirer, ``Predicting first impressions with deep learning,'' in \emph{IEEE
  FG}, 2017, pp. 518--525.

\bibitem{richardwebster2018visual}
B.~RichardWebster, S.~Y. Kwon, C.~Clarizio, S.~E. Anthony, and W.~J. Scheirer,
  ``Visual psychophysics for making face recognition algorithms more
  explainable,'' in \emph{ECCV}, 2018, pp. 252--270.

\bibitem{milford2019self}
M.~Milford, S.~Anthony, and W.~Scheirer, ``Self-driving vehicles: Key technical
  challenges and progress off the road,'' \emph{IEEE Potentials}, vol.~39,
  no.~1, pp. 37--45, 2019.

\bibitem{prins2016psychophysics}
N.~Prins and F.~Kingdom, \emph{Psychophysics: A Practical Introduction (Second
  Edition)}.\hskip 1em plus 0.5em minus 0.4em\relax Academic Press, 2016.

\bibitem{willis2006first}
J.~Willis and A.~Todorov, ``First impressions: Making up your mind after a
  100-ms exposure to a face,'' \emph{Psychological Science}, vol.~17, no.~7,
  pp. 592--598, 2006.

\bibitem{ponce2016chalearn}
V.~Ponce-L{\'o}pez, B.~Chen, M.~Oliu, C.~Corneanu, A.~Clap{\'e}s, I.~Guyon,
  X.~Bar{\'o}, H.~J. Escalante, and S.~Escalera, ``Chalearn {LAP} 2016: First
  round challenge on first impressions-dataset and results,'' in \emph{ECCV
  Workshops}, 2016, pp. 400--418.

\bibitem{webster2018psyphy}
B.~R. Webster, S.~E. Anthony, and W.~J. Scheirer, ``Psyphy: A psychophysics
  driven evaluation framework for visual recognition,'' \emph{IEEE Transactions
  on Pattern Analysis and Machine Intelligence}, vol.~41, no.~9, pp.
  2280--2286, 2018.

\bibitem{behavioralgeirhos2018generalisation}
R.~Geirhos, C.~R. Temme, J.~Rauber, H.~H. Sch{\"u}tt, M.~Bethge, and F.~A.
  Wichmann, ``Generalisation in humans and deep neural networks,''
  \emph{NeurIPS}, 2018.

\bibitem{jang2021noise}
H.~Jang, D.~McCormack, and F.~Tong, ``Noise-trained deep neural networks
  effectively predict human vision and its neural responses to challenging
  images,'' \emph{PLoS biology}, vol.~19, no.~12, p. e3001418, 2021.

\bibitem{sunderhauf2018limits}
N.~S{\"u}nderhauf, O.~Brock, W.~Scheirer, R.~Hadsell, D.~Fox, J.~Leitner,
  B.~Upcroft, P.~Abbeel, W.~Burgard, M.~Milford \emph{et~al.}, ``The limits and
  potentials of deep learning for robotics,'' \emph{The International Journal
  of Robotics Research}, vol.~37, no. 4-5, pp. 405--420, 2018.

\bibitem{lake2015human}
B.~M. Lake, R.~Salakhutdinov, and J.~B. Tenenbaum, ``Human-level concept
  learning through probabilistic program induction,'' \emph{Science}, vol. 350,
  no. 6266, pp. 1332--1338, 2015.

\end{thebibliography}

% \begin{thebibliography}{sample.bib}
\bibliographystyle{IEEEtran}

\end{document}